\documentclass[a4paper, 10pt, conference]{ieeeconf}      

\IEEEoverridecommandlockouts                              

\usepackage{lipsum}
\usepackage{color}
\usepackage[colorinlistoftodos]{todonotes}

\usepackage{amsfonts}
\usepackage{mathtools}

\usepackage{graphicx}
\usepackage{subcaption}

\let\labelindent\relax
\usepackage{enumitem}

\title{\LARGE \bf
Interactive Decision Making for Autonomous Vehicles in Dense Traffic
}

\author{David Isele$^{1}$  
\thanks{$^{1}$David Isele is with Honda Research Institute USA,
        {\tt\small disele@honda-ri.com}}%
}

\begin{document}

\maketitle
\thispagestyle{empty}
\pagestyle{empty}

\begin{abstract}

Dense urban traffic environments can produce situations where accurate prediction and dynamic models are insufficient for successful autonomous vehicle motion planning. We investigate how an autonomous agent can safely negotiate with other traffic participants, enabling the agent to handle potential deadlocks. Specifically we consider merges where the gap between cars is smaller than the size of the ego vehicle. 
We propose a game theoretic framework capable of generating and responding to interactive behaviors. Our main contribution is to show how game-tree decision making can be executed by an autonomous vehicle, including approximations and reasoning that make the tree-search computationally tractable. 
Additionally, to test our model we develop a stochastic rule-based traffic agent capable of generating interactive behaviors that can be used as a benchmark for simulating traffic participants in a crowded merge setting. 

\end{abstract}

\section{Introduction} \label{intro}

\begin{figure}[thpb!]
    \centering
    	\includegraphics[trim={0mm 0mm 0mm 0mm}, clip, height=0.9in]{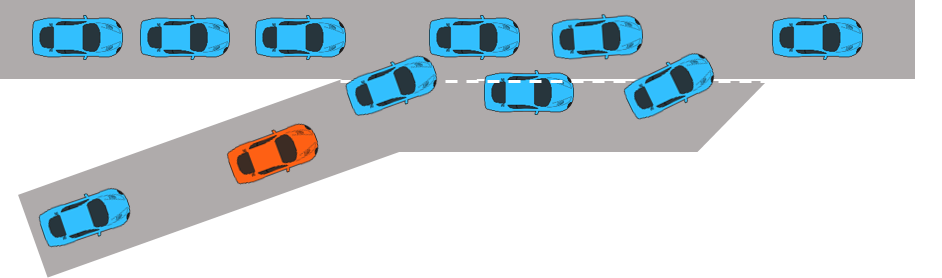}
        \caption{\textbf{Interactive Behavior In Merging.} The ego car (shown in red) needs to interact with the cars in the lane it wishes to enter in order to create an opening. If the ego car were to wait for an opening, it may have to wait indefinitely, greatly frustrating drivers behind it. }\label{fig:title}
        \vspace{-0.3cm}
\end{figure}

Much of the long tail around autonomous driving behavior relates to complex interactions between self-interested agents. Since other traffic participants exhibit a great deal of variety and are often neither purely adversarial, nor purely cooperative, it can be difficult to reason about their behavior. However this type of reasoning is essential to numerous traffic situations in congested traffic such as over-crowded merge scenarios depicted in Figure \ref{fig:title}. In these scenarios, many popular algorithms, such as those based on gap estimation \cite{brilon1999useful}, will be forced to wait indefinitely. 

To address the problem of interacting with self-interested agents, we formulate the problem as a stochastic game \cite{shapley1953stochastic,littman1994markov}. This formulation gives us the ability to reason over multiple outcomes and safely handle agents in an adaptive and online manner. 
Sequential decision making problems, like the crowded merge scenario, are more commonly formulated as Markov Decision Processes (MDPs). While MDPs have been used for robot interaction \cite{melo2011decentralized} and have been the focus of much work in autonomous driving \cite{brechtel2014probabilistic,bai2015intention,ulbrich2013probabilistic}, they  assume agents follow a set distribution which limits an autonomous agent's ability to handle non-stationary agents which change their behavior over time. 

We instead focus our research on searching stochastic game trees.
Stochastic games and the partially observable variant, interactive POMDPs \cite{gmytrasiewicz2005framework}, have both been proposed to enable greater ability to reason about interactions. However this increased power comes at the cost of increased computational expense.
Tree searches, such as Monte-Carlo Tree Search (MCTS) \cite{browne2012survey} and its variants \cite{silver2010monte} have been applied successfully to robotic systems \cite{goldhoorn2014continuous}, and some research has extended tree-based search strategies to handle intentions by modeling them as beliefs \cite{bai2015intention,bouton2017belief}. But these works do not explicitly reason about the other agent's actions which limits the ability to express agent interaction. 


Currently, one of the most popular approaches for solving stochastic games for multi-agent systems involves using deep neural networks (DNNs) and reinforcement learning (RL) \cite{lowe2017multi}. 
However, physical systems, which can be damaged by bad actions, are poorly suited to the exploration of RL and the large sample complexity of DNNs. Training in simulation rarely transfers \cite{tobin2017domain, isele2017transferring}. And transfer is likely to be worse in the case of stochastic games which are known to stereotype to the behavior of the agents they are trained against \cite{lanctot2017unified,foerster2018counterfactual,anahita2019icra}. 
Working with the game trees directly produces interpretable decisions which are better suited to safety guarantees, and ease the debugging of undesirable behavior. However, the intractability of solving a game tree requires approximations to be made. 


Several strategies have been proposed to make solving these problems more efficient. These techniques include making assumptions that restrict branching such as 
using k-level reasoning \cite{camerer2004cognitive}, assuming simplified models of other agents \cite{hoang2013interactive}, and using temporarily extended actions to reduce the required depth of a search \cite{isele2018safe}. 
There are other works in autonomous driving literature which also use game theoretic strategies, and like our work, they adopt some of these approximations. 
However, compared to our work, these approaches either do not allow for the stochasticity 
of the other agents \cite{tian2018adaptive}, or only model intentions and do not have interactive behaviors \cite{wei2013autonomous}.



To solve the games efficiently we adopt existing techniques from the literature and introduce autonomous-driving specific approximations. Additionally, we work through design elements related to mapping merge behaviors to the traditional stochastic game formulation.

In order to validate our behavior we need interactive agents to test against. This produces a `chicken and egg' problem, where we need to have an intelligent agent to develop and test our agent. To address this problem we develop a stochastic rule-based merge behavior which can give the appearance that  agents are changing their mind. Separate to our method, this model may be of interest to other researchers working on merge behaviors. 

 


 
\section{Interactive Decision Making}\label{sec:interactive}

We start by formalizing the interactive decision making process for an autonomous vehicle as a stochastic game. Because game trees grow exponentially in both the number of players and actions, we discuss how we limit the branching factor of both the agent we control (which we call the \emph{ego agent}) and the other traffic participants. Specifically we describe how we use probabilities to reason about self-interested agents, discuss how those probabilities are updated online, and describe how the predicted behaviors of other agents allow us to discretize and limit the ego agent's action space while still maintaining safety in the event that less probable actions are taken by others.

\subsection{Problem Statement and Approach}

In this document we will use the subscript/superscript notation $variable_{time}^{agent,action}$.
In a stochastic game, at time $t$ each agent $i$ in state $s_t$ takes an action $a_t^i$ according to their policy $\pi^i$. All the agents then transition to the state $s_{t+1}$ and receive a reward $r_t^i$. Stochastic games can be described as a tuple $\langle \mathcal{S}, \mathbf{A}, P, \mathbf{R} \rangle$, where $\mathcal{S}$ is the set of states, and $\mathbf{A} = \{\mathcal{A}^1,\dots,\mathcal{A}^n \}$ is the joint action space consisting of the set of each agent's actions, where $n$ is the number of agents. The reward functions 
$\mathbf{R} = \{\mathcal{R}^1,\dots,\mathcal{R}^n \}$ describe the reward for each agent $\mathcal{S} \times \mathbf{A} \rightarrow \mathbf{R}$. The transition function 
\mbox{$P: \mathcal{S} \times \mathbf{A} \times \mathcal{S} \rightarrow [0,1]$} describes how the state evolves in response to all the agents' collective actions.
Stochastic games are an extension to Markov Decision Processes (MDPs) that generalize to multiple agents, each of which has its own policy and reward function. 

Nodes in a game tree represent states, the tree is rooted at the current state $ \mathbf{s}_0 \in \mathcal{S}$, and a branch exists for every possible set of actions. If we assume all cars have the same action space, and there is one ego vehicle and $n-1$ traffic participants, each layer of the game tree has $  |\mathcal{A}|^n = |\mathcal{A}^1|\times \dots \times |\mathcal{A}^n|$ branches. For consistency of representation, this can also be likened to a game tree where players take turns by representing decisions of a single time step as an $n$ layer tree with $|\mathcal{A}|$ branches per node. 
In this stacked representation, it is understood that the $n$ moves are carried out simultaneously. A game tree is depicted in Figure \ref{fig:matrix_game}.


Finding the optimal action sequence for a fixed time horizon $T$ requires taking the rewards associated with the leaf nodes and using dynamic programming to propagate up the tree and select the actions associated with the greatest expected return. Because the number of branches is exponential in the number of agents, brute force implementations to a horizon of depth $T$ result in the doubly exponential runtime of $O\left( |\mathcal{A}|^{n^T} \right)$. 

In autonomous driving, there are hopefully never any pure adversaries (no one is \emph{trying} to crash into our vehicle), and zero-sum planning strategies, like minimax search with alpha-beta pruning \cite{knuth1975analysis}, would be prohibitively cautious. It is more accurate to model other agents as being self interested (they want to travel as fast as possible without crashing), but this does not allow us to predict their actions unless we know their values (how much do they want to avoid crashing, how much do they value being kind to others, how much do they want to hurry). So, to solve the game tree, we turn to probabilities to model the expected behavior of other agents and look for ways to limit the branching while being careful to not \emph{assume} the expected behavior of other agents.

\subsection{Pipeline Overview}\label{sec:pipeline}

The time it takes to complete a maneuver is on the order of several seconds. If we are making control actions at a modest rate of 10Hz, this can result in a game tree with a depth of tens to hundreds of layers. To get around this complexity issue we condense sequences of actions into intentions which follow a distribution \cite{isele2018safe}. Additionally, by using intentions, we are able to discretize our continuous action space.

This reduces the depth of our search, but running forward simulations for all possible ego actions and all possible combinations of other agents' actions is prohibitively expensive, we must also look at reducing the branching factor.

To reduce the number of ego-agent actions, we decompose the actions into a subgoal selection task and a within-subgoal set of actions. Subgoal selection is done using a probabilistic tree search which does not require forward simulation. The smaller set of within-subgoal intentions is then used for forward simulation. 

To reduce the number of traffic participant actions, we select a target interactive agent and then assume level-0 (non-interactive) \cite{camerer2004cognitive} predictions for the other traffic agents. The prediction set of other agents, including all possible intentions for the targeted interactive agent, are then compared against generated samples from the selected ego intention class. 



The decision making pipeline can be broken down as follows:

\begin{itemize}[noitemsep]
\setlength\itemsep{0mm}
\item Select an intention class based on a coarse search 
\item Identify the interactive traffic participant
\item Predict other agents' intentions 
\item Sample and evaluate the ego intentions 
\item Act, observe, and update our probability models
\end{itemize}

A course search gives us the look ahead to break us out of the local minima we might get trapped in from a greedy search. 
A prediction of traffic participant behaviors allows us to reason about which parts of the road are available and when. Typically these predictions will be represented as probability distributions to allow for variations in driver motions. 

Using the predictions of other agent motions, a set of safe intentions is generated, taking into account the possible different ways in which other agents may respond. Each trajectory can then be evaluated against several metrics (risk, efficiency, etc.) and a single trajectory is then selected.  

Next we examine details of each of these components.

\subsection{Intention Class Search}\label{sec:class}

The first step is a coarse tree search. We evaluate the probability of a successful merge $m$ into each potential gap, taking into account the traffic participant's willingness to yield $y$, the size of the gap $g$, and the distance $d$ the gap is from our current position. 
We assume the influences of personality, distance, and gap size on success are independent:   
\begin{eqnarray}
P(m | y, d, g) \propto  
\frac{P( m | y) P(m | d) P(m | g)}{P(m)}     \enspace .
\end{eqnarray}
The personality models $P( m | y)$ which govern the willingness-to-yield estimates are initialized to the same prior and are updated based on how a traffic agent responds to the ego agent's intentions. 

Distance is used as a proxy for time. This allows us to reason about how likely a space is to be available by the time the ego vehicle gets close to gap. 
To model the probability of successfully merging given distance $P(m | d)$ we use a Gaussian
\begin{eqnarray}\label{eq:p_m_d}
P(m | d) \propto \exp\left(-\frac{(d-d_0)^2}{\sigma^2}\right) \enspace . 
\end{eqnarray}
To compute the probability of successfully merging given gap size $P(m | g)$, we normalize gap lengths 
and then the normalized gap $g_{norm}$ is then transformed into a probability using a logistic
\begin{eqnarray}\label{eq:p_m_g}
P(m | g) = \frac{1}{1+exp(-kg_{norm})} \enspace ,
\end{eqnarray}
where $k$ controls the steepness of the slope.

When searching the coarse tree we only consider the probability of success without taking into account the specific motions the other agents may take. This simplifies the problem to a traditional tree search problem and allows us to use higher level planning to avoid local optima. For example, there might be a large opening far away, which has the greatest probability of success, but if it fails, we will have no other options. A greedy search would target this gap, but the Bellman equations might yield a lower probability option with numerous backup plans.


\subsection{Prediction}\label{sec:prediction}
Once a specific gap is targeted, a single agent can be identified with which the ego agent needs to interact. This reduces the branching factor to the number of intentions that the selected agent has available. 

To predict the agent's intentions, our method combines a conventional prediction based on vehicle dynamics and road geometry, and an interactive prediction that looks at how behavior might change in response to ego agent intervention. 
The first component is a common choice for prediction modules in autonomous driving publications \cite{lefevre2014survey}. It derives its predictions from kinematic models of the participants, for example constant velocity assumptions. While clearly a simplification, these predictions work well under many driving conditions. However, under congested settings, for example merging in dense traffic as shown in Figure \ref{fig:title}, these prediction models break down. 
\begin{figure}[thp!]
    \centering
    \vspace{6pt}
    	\includegraphics[trim={0mm 20mm 0mm 0mm}, clip, height=1.6in]{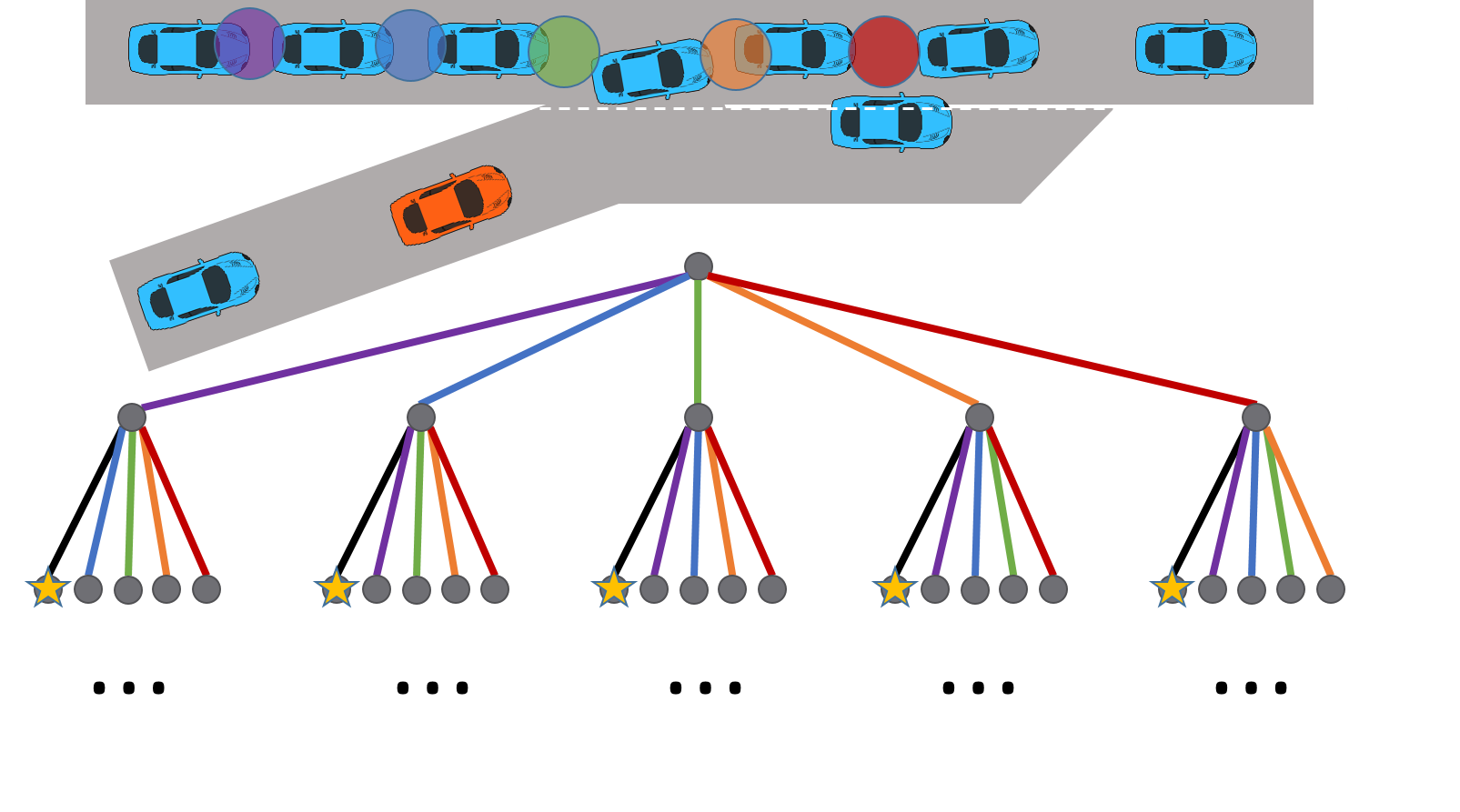}
        \caption{\textbf{Intention Class Search Tree.} 
        Colored circles identify the possible gaps to target. Colored edges of the tree signify the ego agent targeting the corresponding gap. The agent either succeeds (yellow star) or selects a different gap. The search continues continues to a fixed depth or ends once every gap has been attempted.  
        }\label{fig:search_tree}
        \vspace{-0.5cm}
\end{figure}

For the second component, we consider predicting behavior in the presence of multiple agent interactions using counterfactual reasoning \cite{foerster2018counterfactual}. This looks at the predicted behavior in the absence of the ego agent, and then all the change that occurs is credited to the actions of the ego agent. While it is the case that a real driver may brake based on the behavior of a driver three cars ahead of his current position, or may change lanes in spite of his neighbors to avoid a congestion spotted ahead, assuming a single agent as the cause of a behavior is a useful approximation. 

For tractability, we follow counterfactual reasoning. We assume that at any given time our agent interacts with only one other agent. We will then have multiple predictions for this agent based on the believed responses to our latest action. Given the Markov assumption we have:
\begin{eqnarray}
P(a_t^v | a_{t-1}^v, a_{t-1}^e, s_t^v, s_t^e) \propto  
\frac{P(a_t^v  | a_{t-1}^v, s_t^v) P(a_t^v  | s_t^e) }{P(a_t^v)}     \enspace .
\end{eqnarray}


Intuitively, $P(a_t^v  | a_{t-1}^v, s_t^v)$ describes the probability of an agent's action based on it's current state, including the \emph{continuity} or how likely they are to continue doing what they were doing. This corresponds to traditional predictions based on kinematics and road information. $P(a_t^v  | s_t^e)$ describes the \emph{influencibility}, or how likely the agent is to change its behavior based on the state of the ego agent. Rather than learning $P(a_t^v  | s_t^e)$, we use heuristics to model intentions both in the absence of the ego agent and in the presence of the ego agent, \emph{i.e.} \emph{lane following} and \emph{braking} intentions. 


\subsection{Generating Safe Intentions}\label{sec:generate}

\begin{figure*}[thpb!]
    \centering
    \hspace{-10pt}
    	\includegraphics[trim={0mm 0mm 0mm 0mm}, clip, height=2.0in]{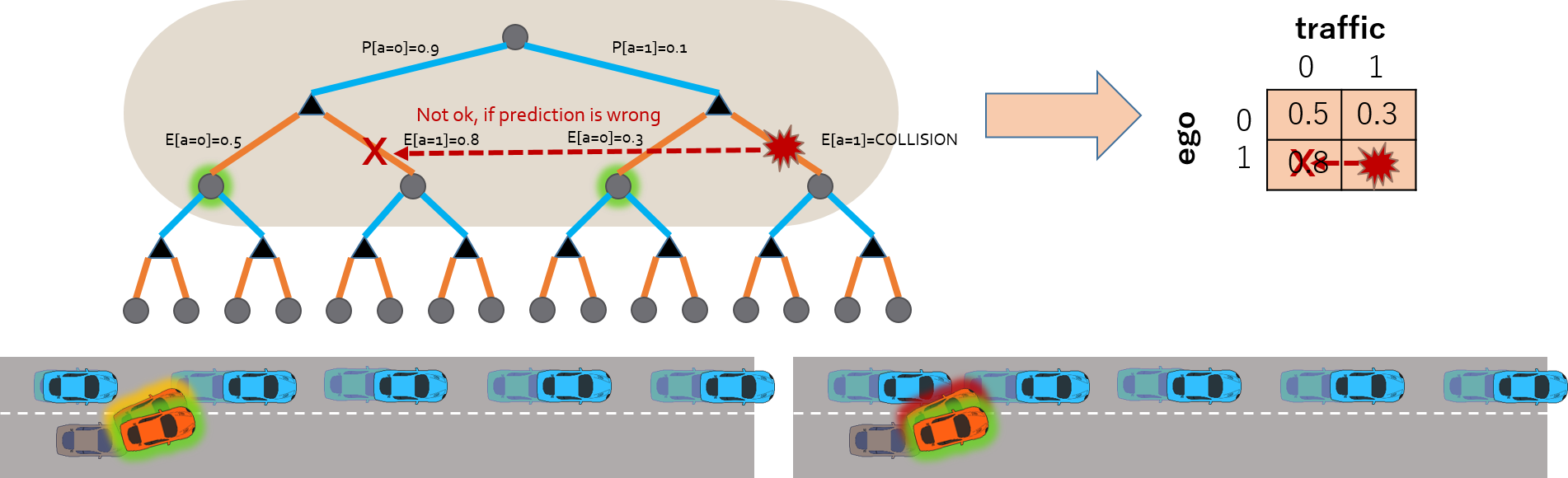}
        \caption{\textbf{Intention Game Tree.} 
        Triangles indicate ego agent intentions, and circles indicate intentions of the agent being interacted with. 0 and 1 denote different discrete intentions. While the more aggressive ego intention (a=1) can give a greater reward in the probable case that the traffic car brakes, if the car does not brake a collision occurs, so this action must be pruned. The actions at each level of the game tree (corresponding to the oval) are equivalent to a matrix game.  
        }\label{fig:matrix_game}
        \vspace{-0.3cm}
\end{figure*}

Given the set of predicted intentions, their corresponding probabilities, and the ego agent's targeted gap location, there exists a substantially reduced space of possible intentions from which the ego agent may choose.  

Each sampled intention has a corresponding risk (based on how close it gets to other vehicles), success (based on how likely it is to achieve its goal), and comfort value (based on the aggressiveness of the maneuvers). 
Since the ego agent does not have control over the other agent's intentions, it must weight expected returns based on the expected probability of the other agent's intentions. 
Importantly, to ensure a safe planning strategy, the ego agent must ensure that there is always a viable (although not necessarily optimal) response to the action the other agent may select. For example the ego agent may attempt to merge, hoping the other agent will yield. But the ego car's attempt must be such that it has a exit strategy (slowing down, and rerouting back to its lane) if the other agent does not yield. In the literature, this reaction time is referred to as the Time-to-Reaction (TTR) \cite{ Hillenbrand2006AMC,tamke2011flexible} and this sets a lower bound on our planning horizon. To fully consider the costs of rerouting, a game tree, as depicted in Figure \ref{fig:matrix_game}, is used. This takes into account all the rerouting actions required to handle the various predictions. In practice we found that the coarse tree described in section \ref{sec:class} was sufficient for long term planning and only one intention depth needed to be considered for the fine-grained search. This reduces the second tree to a matrix game. Note that if there are any responses of the other agent that were not predicted, they may still be dangerous. 


\subsection{Rewards}\label{sec:select}

The stochastic game formulation allows for a single scalar valued reward, however
their are numerous values associated with a single trajectory, \emph{i.e.} risk, success, and comfort. These values are not directly comparable so the most appropriate way to discuss optimality is to consider only values that lie along the pareto frontier. A user preference can then be used to disambiguate values on the frontier.  

Additionally, there is ambiguity related to the success value. This can simply be how close we get to our desired goal, or it can incorporate the likelihood that the ego intention elicits a favorable reaction. If probabilities of the other agent are known, as depicted in 
Figure \ref{fig:matrix_game}, the game tree propagation takes care of the latter. 

\subsection{Update Models}\label{sec:update}

After the ego agent has acted, the system is able to observe the other traffic participants behaviors and use this to update the probabilities described in \ref{sec:class} concerning whether negotiating with the agent will result in success.

The probability of successfully merging given yielding $P( m | y)$ is updated using a difference equation 
\begin{eqnarray}
P( m | y)_t = \alpha P( m | y)_{t-1} + (1-\alpha)o_{t} \enspace,
\end{eqnarray}
where $o$ is the observation of whether the car is observed to be yielding or not, and $\alpha$ is a constant which sets the update rate. 
The next section describes the agents we use to evaluate our interactive decision making process. 

\section{Interactive Agents Model}\label{sec:model}
\begin{figure*}[thpb!]
    \centering
    \hspace{-10pt}
    	\includegraphics[trim={0mm 1mm 0mm 1mm}, clip, height=0.95in]{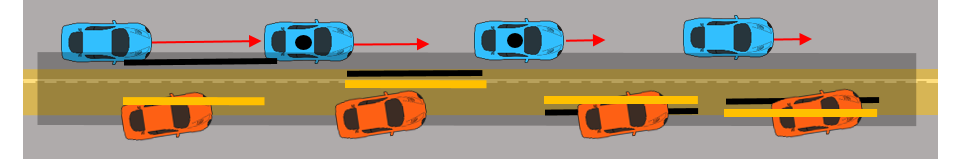}
        \caption{\textbf{Interactive Agent Behavior Model.} The diagram indicates how four cars with different thresholds will all respond to a car in the same relative position with different behavior. The yellow line indicates the lateral position at which the traffic vehicle begins to react to the ego car. The black line indicates when the traffic car yields to let the ego car merge. The yellow and gray shaded regions indicate the distributions from which the thresholds are sampled. If the traffic car is reacting but not yet yielding, it exhibits aggressive behavior: moving forward to block the other agent from merging. The first car (left most) causes a reaction where the driver accelerates to block the merge. The second car does not yet react to the ego car, continuing to drive as usual. The third car yields to the ego car. The fourth car had been moving to block the ego car at first, but then decided to back off and let the ego car merge. }\label{fig:behavior}
        \vspace{-0.3cm}
\end{figure*}
The agent we are trying to develop needs other intelligent agents to interact with when evaluating our algorithm. This presents a ``chicken and egg'' problem, since having such agents available would suggest we already have intelligent agents at our disposal. To solve this problem, we design a rule-based stochastic model that produces behaviors that we find reasonable, varied, and capable of changing in response to the ego agent. Note that the model is unknown to the ego agent, and the ego agent's behaviors are agnostic to the model we use.

Under non-merge conditions, traffic participants follow the intelligent driver model (IDM) \cite{idm-def}. When the ego car is aligned between two cars in the traveling direction, the rear car may change its behavior as a result of the ego car's actions. The way the traffic car behaves is based on two randomized thresholds that govern the agent's behavior. One threshold governs whether or not the agent reacts to the ego car, the second threshold determines how the agent reacts.

Figure \ref{fig:behavior} illustrates how different thresholds produce different behaviors in response to an ego car in the same relative position. 
This process can be viewed as a rule based variant of negotiation strategies \cite{bulling2014survey}: an agent proposes he go first by making it more dangerous for the other, the other agent accepts by backing off.

\section{Experiments} \label{sec:experiments}
In our experiments, cars have a length of $4.8m$ and a width of $1.9m$. 
The lane width is $3.7m$. 
Traffic is generated with a mean gap width between cars corresponding to $\{2.4, 4.8, 9.6\}m$ with added noise drawn uniformly from the range $\pm \{0.4, 0.8, 1.6\}m$ respectively.

As described in Equation \ref{eq:p_m_d},
The probability of successfully merging given distance $P(m | d)$ is assumed to be proportional to a Gaussian. In our experiments, we use 
$\sigma=25$ and an additional penalty of $6m$ to cars in front of the ego car to prevent overly aggressive cut in maneuvers.  
We set the steepness constant in Equation \ref{eq:p_m_g} as $k=0.3$.
The probability of successfully merging given yielding $P( m | y)$ is initialized to $0.5$ for all cars and updated as described is section \ref{sec:update}. 
We use a bicycle model for our vehicle dynamics following \cite{kong2015kinematic}. Our search tree searches to a fixed depth. Intentions are set to a fixed time of 2 seconds and the system replans every second. The action with the max L1 reward vector is selected.  

We perform an ablation study to examine the importance of tree depth to the interactive planning procedure. We run $20$ simulations varying the tree depth from one to three layers deep. Each experiment is initialized with the same random seed and the agents encounter identical situations. The entire experiment is then repeated multiple times with different random seeds to generate confidence intervals related to the observed change. This is done because there are large variances on trial times, but the average stays roughly constant. 

The agent behavior model has two tunable parameters. We sample aggression thresholds for each traffic agent uniformly from the range $\{-2.2, 1.1\}m$ and reaction thresholds from the range $\{-1.5, 0.4 \}m$ where $0$ corresponds to the lane marking.
We also record the runtimes associated various parts of our algorithm and use these numbers to discuss the computation benefits of various aspects of our pipeline. 


\section{Results}\label{sec:results}
\subsection{Depth Ablation}
\begin{figure}[t!]
    \centering
    \vspace{5pt}
    \hspace{-140pt}
    \begin{subfigure}[b]{1.1in}
    	\includegraphics[trim={25mm 55mm 17mm 55mm}, clip, height=0.3in]{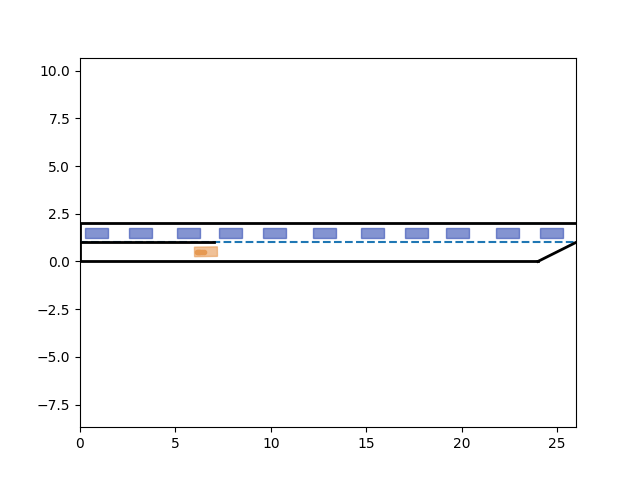}
        \label{fig:scenarios_right}
    \end{subfigure}
    \\
    \hspace{-140pt}
    \begin{subfigure}[b]{1.1in}
    	\includegraphics[trim={25mm 55mm 17mm 55mm}, clip, height=0.3in]{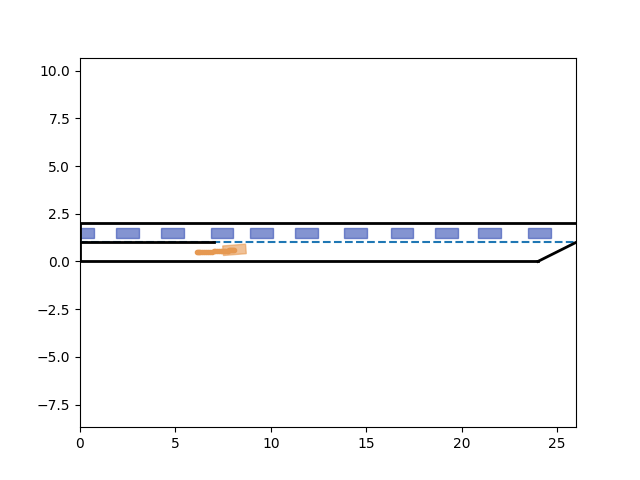}
        \label{fig:scenarios_left}
    \end{subfigure}
    \\
    \hspace{-140pt}
    \begin{subfigure}[b]{1.1in}
    	\includegraphics[trim={25mm 55mm 17mm 55mm}, clip, height=0.3in]{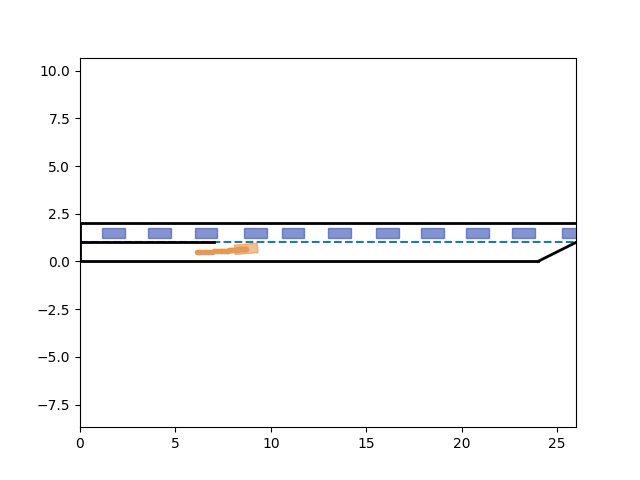}
        \label{fig:scenarios_left2}
    \end{subfigure}
    \\
    \hspace{-140pt}
    \begin{subfigure}[b]{1.1in} 
       \includegraphics[trim={25mm 55mm 17mm 55mm}, clip, height=0.3in]{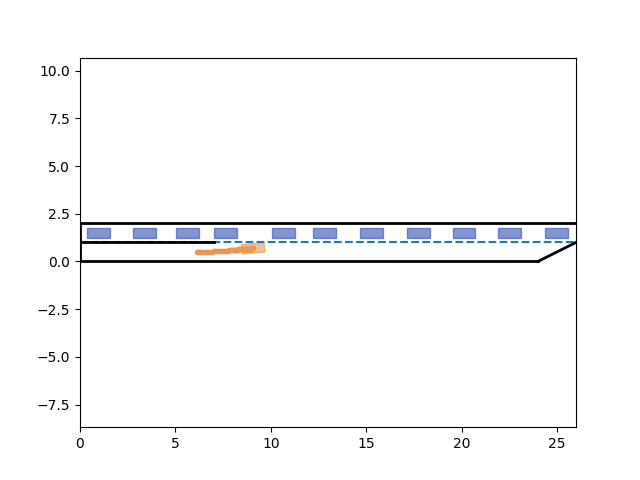}
        \label{fig:scenarios_forward}
    \end{subfigure}
    \\
    \hspace{-140pt}
    \begin{subfigure}[b]{1.1in}
    	\includegraphics[trim={25mm 55mm 17mm 55mm}, clip, height=0.3in]{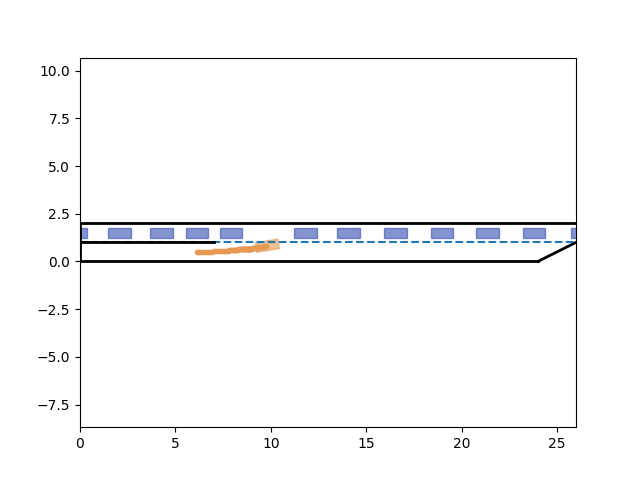}
        \vspace{-10pt}
        \label{fig:scenarios_challenge}
    \end{subfigure}
        \caption{Visualizations of intersection tasks used for our experiments. The first merge attempt is blocked by the traffic vehicle, requiring the ego car to wait for the next opening. }\label{fig:scenarios} 
\end{figure}

A sample run with a $4.8m$ gap is depicted in Figure \ref{fig:scenarios}.
We observe that including the second layer of the tree search slightly reduces the merge time. As depicted in Table \ref{table:depth}, the greatest benefit occurs in the most dense traffic (\emph{2.4m gap}) where the average time decreases by $0.8s \pm 0.3$. For a gap of $4.8m$ the time decreases from 11.2s to 10.9s. This corresponds to a change of $0.3s \pm 0.1$. In sparser traffic (\emph{9.6m gap}) the merge times converge to within the error margins. Adding a third layer to the tree search produces identical results to a two layer search. We believe that this is because there is too much unknown information to benefit from longer horizon planning. 
\begin{table}[htbp!]
\caption{Effect of Depth and Traffic Density on Time}
\begin{center}
\begin{tabular}{c|ccc}
    Depth           &         & Time	 & \\ 
    & 2.4m gap & 4.8m gap & 9.6m gap \\
    \hline
    \emph{1} 
    & 15.2s	    & 11.2s & 4.5s   \\
    \emph{2} 
    & 14.4s	    & 10.9s  & 4.5s   \\
    \emph{3} 
    & 14.4s	    & 10.9s  & 4.5s  \\
\end{tabular}
\end{center}
\label{table:depth}
\vspace{-0.3cm}
\end{table}
Qualitatively inspecting the results, we observe that the behavior is identical for most trials. When there is a difference between the depth 1 and depth 2 trials, it typically corresponds to cases where failing is likely but recovery cost is low. The single depth search goes with the safer decision, the deeper search is slightly more aggressive, usually allowing it to default to the opening selected by depth 1  if it fails.

\subsection{Runtime Analysis}
Next we discuss runtime analysis based on our empirical findings. The gap selection tree search only requires the maintenance and multiplication of probabilities and is comparatively light weight in contrast to the full game tree calculation which requires forward simulation and collision checking for multiple ego and traffic intentions. Running unoptimized python code on an Intel Xeon processor, we can calculate the results of a gap selection tree of depth two in approximately 0.01 seconds. This increases to 0.02 seconds for depth three. The matrix game calculation for a single layer, sampling four ego intentions and using two traffic intentions (brake and maintain), takes roughly 0.08 seconds. This means our entire decision making loop runs at roughly 10Hz. 

If we remove the gap selection process and instead run the forward simulations for all gaps, we will increase the matrix game computations by a factor of 10 (there are usually 11 traffic vehicles on the screen at a time with a $4.8m$ gap width). This results in 0.8 seconds for a single layer, roughly 
one minute for two layers. 

If instead every car is an interactive agent, this increases the traffic intentions from $2$ to $2^{10}=1024$ which will increase computation by a factor of $512$ for the first layer. 

\section{Conclusion}\label{sec:conclusion}

We present a game theory based process for interactive decision making for autonomous vehicles. To reduce the computational complexity to run in real-time, we make several approximations related to safely speeding up the game tree search. While this approach demonstrates a successful method for interactive merging in simulation, there is more investigation required to run safely on an actual vehicle. Namely, we suspect larger variation in real world scenarios. The uncertainty related to both the perception and prediction will need to be considered more deeply and there will likely be latencies in the control which require more cautious behavior sampling, or greater loop optimization.  






\medskip

\bibliographystyle{IEEEtran}
\begin{small}
\bibliography{refs}
\end{small}

\end{document}